\documentclass{article}
\usepackage{spconf,amsmath,epsfig}
\usepackage{hyperref}
\usepackage{color}
\usepackage{balance}
\usepackage{microtype}
\usepackage{float}
\usepackage{comment}

\title{MONITORING WATER CONTAMINANTS IN
COASTAL AREAS THROUGH \\ML  ALGORITHMS LEVERAGING ATMOSPHERICALLY CORRECTED SENTINEL-2 DATA}
\name{\begin{tabular}{c}Francesca Razzano$^{a,b,1}$, Francesco Mauro$^{a}$,  Pietro Di Stasio$^{a}$, Gabriele Meoni$^{c}$,\\
\textit{Marco Esposito$^{d}$, Gilda Schirinzi}$^{b}$, and Silvia Liberata Ullo$^{a}$ \end{tabular}\thanks{
$^{1}$Corresponding author. 
\textit{Email addresses}: francesca.razzano002$@$studenti.uniparthenope.it (FR), f.mauro$@$studenti.unisannio.it (FM), pietro.distasio$@$studenti.unisannio.it (PD), G.Meoni$@$tudelft.nl (GM), m.esposito$@$cosine.nl (ME), gilda.schirinzi$@$uniparthenope.it (GS), ullo$@$unisannio.it (SLU) }
}

\address{
$^{a}$ University of Sannio, Engineering Department, Benevento, Italy \\
$^{b}$ University of Parthenope, Engineering Department, Naples, Italy \\
$^{c}$ Delft University of Technology, Space Engineering Department, Delft, Netherlands\\
$^{d}$ Cosine, Engineering Department, Sassenheim, Netherlands
}

\begin{document}
%
\maketitle
\begin{abstract}
Monitoring water contaminants is of paramount importance, ensuring public health and environmental well-being. Turbid\nobreak ity, a key parameter, poses a significant problem, affecting water quality. Its accurate assessment is crucial for safeguarding ecosystems and human consumption, demanding meticulous attention and action. For this, our study pioneers a novel approach to monitor the Turbidity contaminant, integrating CatBoost Machine Learning (ML) with high-resolution data from Sentinel-2 Level-2A. Traditional methods are labor-intensive while CatBoost offers an efficient solution, excelling in predictive accuracy. Leveraging atmospherically corrected Sentinel-2 data through the Google Earth Engine (GEE), our study contributes to scalable and precise Turbidity monitoring.
A specific tabular dataset derived from Hong Kong contaminants monitoring stations enriches our study, providing region-specific insights. 
Results showcase the viability of this integrated approach, laying the foundation for adopting advanced techniques in global water quality management.
\end{abstract}
\begin{keywords}
Remote Sensing, Artificial Intelligence, Water
contaminants monitoring, Machine Learning
\end{keywords}
\section{Introduction}
\label{sec:intro}
Numerous studies are highlighting the importance of monitor\nobreak ing contaminants in water basins, coasts, or reservoirs. Paramount in this context is water security and access, explicitly acknowledged in the \href{https://www.who.int/health-topics/water-sanitation-and-hygiene-wash#tab=tab_1}{Millennium Development Goals}, with a particular emphasis on the promotion of water resource sustainability.  In this regard in our study, the aim is to monitor the Turbidity parameter, addressing significant negative impacts on human health, as highlighted by the 
\href{https://www.epa.gov/awma/factsheets-water-quality-parameters}{U.S. Environmental Protection Agency (EPA)}. Elevated Turbidity or suspended solids levels can adversely impact aquatic health reducing light penetration. Turbidity acts as a shield for disease-causing pathogens, and it may lead to waterborne disease, which can cause intestinal sickness. 
Turbidity monitoring is essential, but traditional in-situ measurements, while precise, face challenges in creating regional-scale water quality databases due to their labor-intensive nature and time constraints. 
Implementing Remote Sensing (RS) techniques allows for a more expansive and efficient assessment of water quality on a larger scale, overcoming the limitations associated with traditional techniques.
There are some interesting investigations useful in this end  \cite{toming2016first, potes2018use}
and, in addition, the evolution of (Machine Learning) ML models and their integration into RS have introduced a novel dimension to water quality metrics \cite{sharaf2017mapping, guo2021machine}.\\
Our study aims to pioneer a novel approach for monitoring Turbidity contaminants using a combination of CatBoost ML and high-resolution data from Sentinel-2 Level-2A, atmospherically corrected Sentinel-2 data. 
Our research endeavors to refine the existing methodology outlined in the state of the art by employing the CatBoost Machine Learning technique for regression. Motivated by a seminal study \cite{hkong_stations}, we seek to elevate current practices, specifically addressing the use of Level-1C images in the Hong Kong article, which represent a distinct Sentinel-2 product. These images, while not atmospherically corrected initially, undergo several pre-processing steps that require other specific settings. Our innovative solution involves the integration of Sentinel-2 Level-2A images downloaded from GEE and incorporating atmospheric correction to optimize analysis phases and enhance the overall proposed methodology. The implementation of a highly effective ML technique, coupled with an increased number of training samples, significantly contributes to more comprehensive network training. As a result, our case study demonstrates superior metrics when compared to the samples reported in the Hong Kong article, which utilizes an Artificial Neural Network (ANN) with specific settings. This underlines the efficacy of our approach in addressing and surpassing the limitations identified in the prior study.\\
Our work discusses key aspects, such as the use of the atmospherically corrected Sentinel-2 data via GEE platform, the description of the Area of Interest (AOI), the creation of a tabular dataset derived from the Hong monitoring stations \cite{hkong_stations}, and the application of the CatBoost ML method developed through Python language. The final sections present the obtained results and the conclusions.\\
\vspace{-0.7cm}
\section{ GEE
ENVIRONMENT AND DATA SOURCES}
\label{sec:gee}
In the following sections, the significance of the GEE platform is addressed with the chosen dataset for the specific scenario.
\vspace{-0.2cm}
\subsection{GEE environment}
The \href{https://earthengine.google.com/}{GEE} platform is a powerful environment, that has been meticulously designed for the visualization of petabyte-scale geographic data and to facilitate advanced scientific research. It offers an extensive collection of data characterized by diverse bands, projections, and resolutions, along with pivotal tools for enabling swift multisensor analysis. Its inherent capabilities empower users to employ both fundamental and sophisticated techniques, including ML-based algorithms and models, for the comprehensive analysis of Earth Observation (EO) data.
\vspace{-0.2cm}
\subsection{Sentinel-2 Level-2A}
From GEE  Sentinel-2 Level-2A images have been retrieved and a dedicated developed, as described in Section 
 \ref{creation_sub}. 
In RS applications within GEE, the utilization of Sentinel-2 \href{https://developers.google.com/earth-engine/datasets/catalog/COPERNICUS_S2_SR#description}{Level-2A products}  holds particular prominence,
since they are suitable for various applications, including land cover mapping, vegetation monitoring, and change detection. 
For completeness, it is emphasized that although 13 spectral bands are made available by the advanced MultiSpectral Instrument (MSI) sensor on board Sentinel-2, Band 10 is not present in the Sentinel-2 Level-2A products.
\vspace{-0.5cm}
\section{AREA OF INTEREST}
\label{sec:area}
\begin{figure}[ht!]
	\centering
     \includegraphics[scale=0.54]{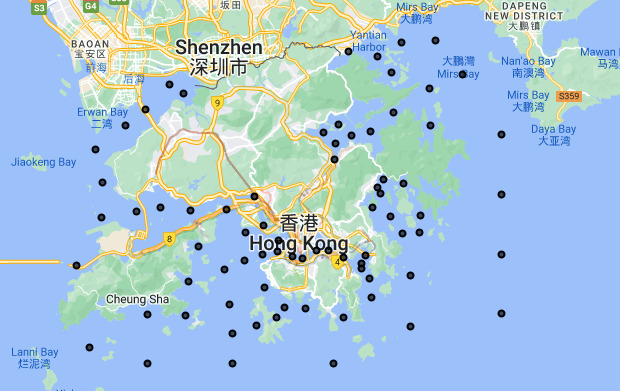}
	\caption{Map of the 76 monitoring stations in coastal areas and open sea of Hong Kong region - the Area of Interest (AOI)}
	\label{station}
\end{figure}

\vspace{-0.3cm}
The AOI of our study has been identified with the Hong Kong region motivated by the initial study described in \cite{hkong_stations}. 
The coastal waters of Hong Kong are physically and chemically complex because of a mix of anthropogenic activities and fluctuating hydrographic conditions. 
Based on the hydrodynamic properties and pollution status, the Environmental Protection Department (EPD) of the Hong Kong government separated Hong Kong waterways into 10 water management zones. Every month, EPD uses a specialized marine monitoring vessel outfitted with a Differential Global Positioning System (DGPS) and an advanced Conductivity, Temperature, and Depth (CTD) profiler to analyze water quality and collect samples at 76 monitoring sites in coastal areas and open sea, as shown in the map of Fig. \ref{station}, created to identify the distribution of the monitoring stations. 
A specific \href{https://data.gov.hk/}{database} collects all water quality data and has been made available online. 
This database
has been instrumental in our study to determine the Ground Truth (GT) for the Turbidity parameter that we want to monitor.

\section{DATASET CREATION AND METHODOLOGY}
\label{sec:creation}
In our case study, we decided to employ a CatBoost as an ML model for the regression analysis, to predict the values of contaminants. This decision is grounded in a comprehensive examination of the State of the Art (SOTA), which indicates that CatBoost, along with many ML techniques, demonstrates superior performance compared to DL models \cite{10313279, prokhorenkova2018catboost} in the contest of interest.
For our regression model, a set of independent variables as input to the model is needed. In this case, the spectral bands of the Sentinel-2 Level-2A have been selected to this end. These variables are believed to have an impact on the output variables we aim to predict, namely the target or dependent variables representing the values of the contaminants. Independent variables can be taken of various types, such as numeric, categorical, or others.
CatBoost is especially well-suited for handling different types of variables without the need for specific conversion, thereby simplifying the data preparation process. 
\vspace{-0.2cm}
\subsection{Dataset creation} \label{creation_sub}
\begin{figure}[ht!]
	\centering
     \includegraphics[scale=0.52]{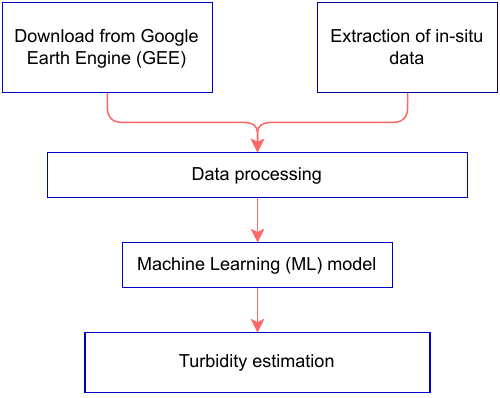}
	\caption{Workflow of the proposed work}
	\label{flow_work}
\end{figure}

To derive the values at the different bands and therefore the independent variables of our regression problem, a dedicated code was developed to retrieve Sentinel-2 Level-2A images from GEE. Based on the method proposed in the previous paper by Mauro et al. \cite{10282352}, where an automated methodology for downloading satellite data from GEE was introduced, a modified version of the code has been adapted for the analysis at hand. Furthermore, all bands have been resampled to a spatial resolution of 10 meters, and the code has been entirely automated. This enables effortless extension of the dataset to other regions in the world.
An array containing latitude and longitude coordinates for the points of interest is used, as well as an array of dates corresponding to each point. For each pair of geographic coordinates and dates, a pre-processing operation calculating a time window of three days before and after the in-situ date is performed and a square of size 0.2 km x 0.2 km is created around the central point specified by each pair of latitude and longitude coordinates. Satellite images from the Level-2A Sentinel-2 dataset (’COPERNICUS/S2$\_$SR’) within this geographical area and for the specified time window are filtered. The bands considered in this regard are: B1, B2, B3, B4, B5, B6, B7, B8, B8A, B9, B11, B12, at the different spatial resolution of 10, 20 and 60 meters.
The final images are in total 660 and their peculiarity is to be all atmospherically corrected. At this stage, to each image, along with its corresponding bands (representing the independent variables of the regression problem), is assigned the value of the dependent variable, constituted by the GT contaminant value extracted from the \href{https://data.gov.hk/}{database} of Hong Kong. 
Subsequently, the constructed dataset is in tabular form, obtained in this way through the calculation of the average over the small squares, and with this dataset, the specific ML algorithm is trained to learn the associated contaminant values. In Figure \ref{flow_work} is reported the workflow of our work. Dataset and code are available on a \href{https://github.com/Fraapp24/Turbidity_monitoring_ML_model}{specific GitHub page} that will be made publicly available after paper acceptance 

\vspace{-0.5cm}
\subsection{The ML algorithm: CatBoost}
To monitor the Turbidity in coastal areas, we chose as ML algorithm, the \href{https://towardsdatascience.com/catboost-regression-in-6-minutes-3487f3e5b329}{CatBoost},  
pioneered by \href{https://neptune.ai/blog/when-to-choose-catboost-over-xgboost-or-lightgbm}{Yandex}
and extending upon the principles of decision trees and gradient boosting. 
Numerous studies have affirmed the efficacy of CatBoost as an ML algorithm \cite{10313279, 10170600,  inproceedings}.
In our work, a \href{https://pythoninstitute.org/about-python#:~:text=Python%20was%20created%20by%20Guido,called%20Monty%20Python's%20Flying%20Circus.}{Python} library has been utilized to implement the CatBoost model for regression task and new code has been developed to tailor the model to our case study.  
Since it represents a high-level, versatile, and interpreted programming language known for its readability and simplicity.
\subsubsection{Methodology applied}
As previously described, the model was trained with input-output pairs, specifically the independent and dependent variables from the created dataset. In this way, the model endeavors to learn the relationship between the features extracted from satellite images and the levels of water contaminants. In the prediction phase, after the training process, the model is employed to forecast water contaminant levels on new data. During this stage, only the independent variables are supplied in input. Furthermore, the data are systematically prepared, incorporating a suitable splitting between training, testing, and validation sets with percentages of 55\%, 20\% and 25\% respectively. Lastly, the model performance is assessed using appropriate regression metrics, such as the Mean Squared Error (MSE), Root Mean Squared Error (RMSE) and Mean Absolute Error (MAE) values. These metrics are typically employed for regression problems, and particularly RMSE and MAE are utilized in the referenced work \cite{hkong_stations} and this facilitates a comparative analysis.
The selection of parameters for the CatBoost algorithm is reported in Table \ref{tabel_parameters}
\begin{table}[ht!]

\caption{Parameters for CatBoost}
\vspace{4pt}
    \centering
    \resizebox{1\columnwidth}{!}{
    \begin{tabular}{|c|c|c|c|c|}
     \hline
     Iterations & Learning\_rate &  depth & Loss function & L2\_leaf\_regularization\\
     \hline
     600    & 0.6 & 12 & RMSE & 1.0 \\
     \hline
\end{tabular}
\vspace{2pt}
    \label{tabel_parameters}

    }    
\end{table}
where the 'depth' parameter manages the depth of the tree structure, while the 'Learning\_rate' influences the size of adjustments applied to the tree model, determining the speed at which the model learns. The 'Iterations' parameter corresponds to the number of trees (rounds), emphasizing the total number of boosting iterations, and the 'l2\_leaf\_regularization’ represents the \href{https://stats.stackexchange.com/questions/422400/l2-regularization-in-catboost}{L2 regularization coefficient}
to 
prevent overfitting.
In summary, these parameters were tuned to control learning speed and overfitting. Lastly, the chosen Loss function has been the RMSE, a common choice for regression tasks. 
\begin{figure*}[!ht]
    \centering
    \includegraphics[width=1.6\columnwidth]{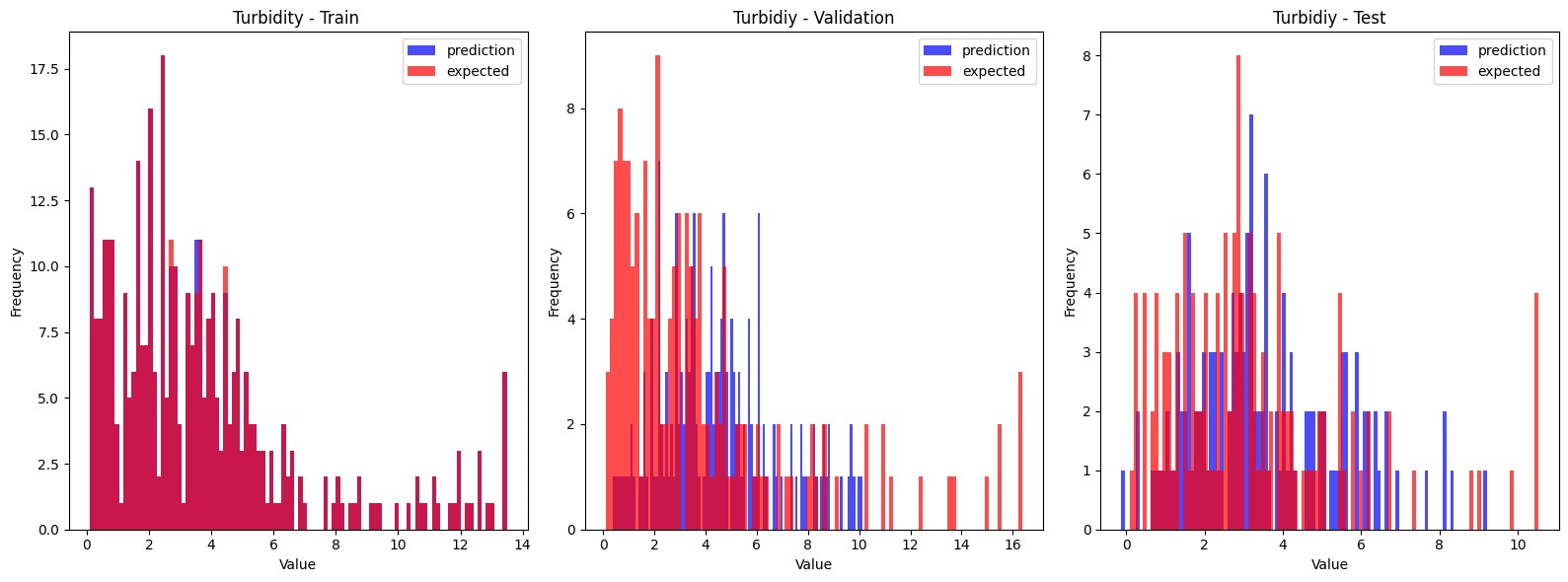}
    \caption{Comparison of expected (red) and predicted (blue) values for Training, Validation, and Testing sets}
    \label{comparison}
\end{figure*}
\section{RESULTS}
\label{sec:results}
From the distribution of the 660 Turbidity values extracted from the Hong Kong database for the selected period and area the mean of the created dataset values is 3.6882 NTU (Nephelometric Turbidity Unit).
\noindent To evaluate the performance of the CatBoost model, we focused on MSE, RMSE, and MAE values. The low metrics reported in Table \ref{tabel_metrics_MSE} suggests that the model has learned the patterns effectively. Even though the RMSE and MAE have higher numerical values compared to the MSE, these metrics demonstrate good performance relative to the mean value. The low metrics' values affirm CatBoost's excellent performance in capturing intricate dataset relationships, making it a promising tool for allowing water contaminants monitoring through RS.

\begin{table}[ht!]

\caption{Metrics for the evaluation of CatBoost performance}
\vspace{4pt}
    \centering
    \resizebox{0.7\columnwidth}{!}{
    \begin{tabular}{|c|c|c|c|}
     \hline
     Metric & Train set & Validation set & Test set\\
     \hline
     MSE & 2.22e-05 & 0.0593215 & 0.0453542\\
    \hline
     RMSE & 0.0047114 & 0.2435600 & 0.2129653\\
    \hline
     MAE & 0.2366917  & 0.2021956 & 0.1945048\\
    \hline
    \end{tabular}
\vspace{2pt}
    \label{tabel_metrics_MSE}   
    }
\end{table}


\begin{figure*}[H]
    \centering
    \includegraphics[width=1.3\columnwidth]{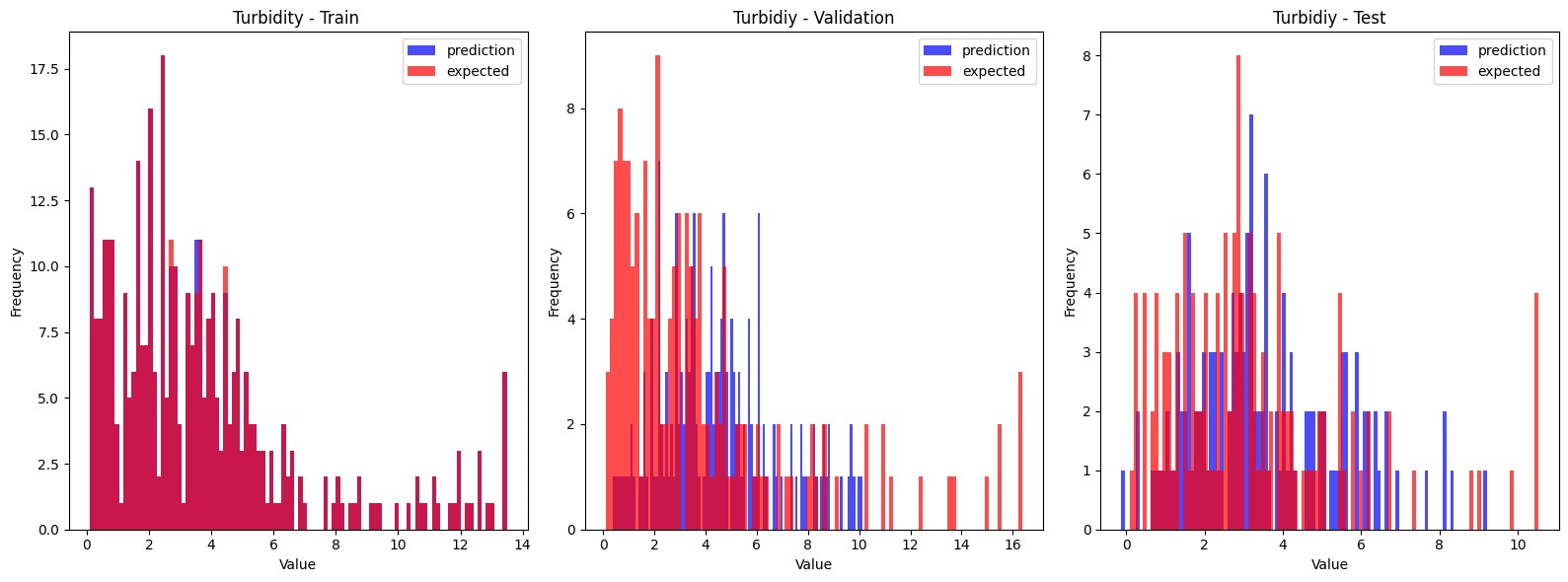}
    \caption{Comparison of expected (red) and predicted (blue) values for Training, Validation, and Testing sets}
    \label{comparison}
\end{figure*}

\noindent Besides the quantitative analysis, it may be beneficial to visually inspect the model predictions against the actual values for identifying specific patterns. As can be observed from the representation in Figure \ref{comparison}, the distributions of the expected and predicted data in the training phase are comparable and practically overlapping. This indicates effective learning from the training set. If we examine the distributions on the validation and test sets, it can be verified that the model reasonably learns and predicts new data. This implies a promising starting point for enhancing the input dataset for the model, thereby enriching its predictive capability. \\
To further validate our study, we conducted a comparison for Turbidity contaminant, reported in Table \ref{tabel_metrics_RMSE_authors_hongKong}, with the metrics used to evaluate the Artificial Neural Network (ANN) proposed by \cite{hkong_stations} and \cite{rs11060617}. We trained the network on a common region, with our focus limited to a sub-area within the broader region they considered. Given that, since in the mentioned studies, the dataset was split into training and validation sets while in our study the splitting involves training, validation, and test sets, to ensure a comparable comparison, the Table reflects accuracies of RMSE and MAE exclusively originating from the validation sets while the Sample size parameter integrates both the phases of training and validation. Furthermore, considering the mean values of the authors, which are 4.80 NTU for \cite{hkong_stations} and 9.4 NTU for \cite{rs11060617}, our mean, when compared to the metrics we obtained, lead us to infer that we have achieved superior results in terms of both RMSE and MAE.

\vspace{-0.3cm}
\begin{table}[ht!]

\caption{
Comparison of Metrics Modeling results of Turbidity water quality parameter}
\vspace{4pt}
    \centering
    \resizebox{1\columnwidth}{!}{
    \begin{tabular}{|c|c|c|c|c|c|}
     \hline
     Study & Years of Interest & Image data & Sample size & RMSE & MAE \\
     \hline
     \cite{rs11060617} & 1999–2015 & Landsat-5, 7, 8 & 120 & 3.10 & 2.61\\
    \hline
    \cite{hkong_stations} & 2015–2021 & Sentinel-2 L1C & 352 & 1.95 & 1.61\\
    \hline
    \textbf{Our study}& \textbf{2019-2020} & \textbf{Sentinel-2 L2A} & \textbf{528} & \textbf{0.24} & \textbf{0.20}\\
    \hline
\end{tabular}
\vspace{2pt}
    \label{tabel_metrics_RMSE_authors_hongKong}   
    }
\end{table}

\vspace{-0.6cm}
\section{CONCLUSIONS}
\label{ssec:subhead}
In conclusion, the results affirm CatBoost's excellent performance in capturing intricate dataset relationships, making it a promising tool for advancing water contaminants monitoring through RS. This study serves as a foundation for future development, enriching global coastal data for adaptable ML models. The network's ability to predict coastal parameters, especially in challenging regions, positions it as a significant asset. Future efforts will extend monitoring to include chemical contaminants, highlighting the potential of ML for coastal ecosystem preservation and global well-being.

\balance
\footnotesize
\bibliographystyle{IEEEbib}
\bibliography{strings,refs}

\begin{thebibliography}{10}

\bibitem{toming2016first}
K.~Toming, T.~Kutser, A.~Laas, M.~Sepp, B.~Paavel, and T.~N{\~o}ges,
\newblock ``{First experiences in mapping lake water quality parameters with Sentinel-2 MSI imagery},''
\newblock {\em Remote Sensing}, vol. 8, no. 8, pp. 640, 2016.

\bibitem{potes2018use}
M.~Potes, G.~Rodrigues, A.M. Penha, M.H. Novais, M.J. Costa, R.~Salgado, and M.M. Morais,
\newblock ``{Use of Sentinel 2--MSI for water quality monitoring at Alqueva reservoir, Portugal},''
\newblock {\em Proceedings of the International Association of Hydrological Sciences}, vol. 380, pp. 73--79, 2018.

\bibitem{sharaf2017mapping}
E.~Sharaf El~Din, Y.~Zhang, and A.~Suliman,
\newblock ``{Mapping concentrations of surface water quality parameters using a novel remote sensing and artificial intelligence framework},''
\newblock {\em International Journal of Remote Sensing}, vol. 38, pp. 1023--1042, 02 2017.

\bibitem{guo2021machine}
H.~Guo, J.J. Huang, B.~Chen, X.~Guo, and V.P. Singh,
\newblock ``{A machine learning-based strategy for estimating non-optically active water quality parameters using Sentinel-2 imagery},''
\newblock {\em International Journal of Remote Sensing}, vol. 42, no. 5, pp. 1841--1866, 2021.

\bibitem{hkong_stations}
I.~Kwong, F.~Wong, and T.~Fung,
\newblock ``{Automatic Mapping and Monitoring of Marine Water Quality Parameters in Hong Kong Using Sentinel-2 Image Time-Series and Google Earth Engine Cloud Computing},''
\newblock {\em Frontiers in Marine Science}, vol. 9, pp. 871470, 05 2022.

\bibitem{10313279}
S.~Berkani, I.~Gryech, M.~Ghogho, B.~Guermah, and A.~Kobbane,
\newblock ``{Data Driven Forecasting Models for Urban Air Pollution: MoreAir Case Study},''
\newblock {\em IEEE Access}, vol. 11, pp. 133131--133142, 2023.

\bibitem{prokhorenkova2018catboost}
L.~Prokhorenkova, G.~Gusev, A.~Vorobev, A.V. Dorogush, and A.~Gulin,
\newblock ``{CatBoost: unbiased boosting with categorical features},''
\newblock {\em Advances in neural information processing systems}, vol. 31, 2018.

\bibitem{10282352}
F.~Mauro, B.~Rich, V.W. Muriga, F.~Janku, A.~Sebastianelli, and S.L. Ullo,
\newblock ``{SEN2DWATER: A Novel Multispectral and Multitemporal Dataset and Deep Learning Benchmark for Water Resources Analysis},''
\newblock in {\em IGARSS 2023 - 2023 IEEE International Geoscience and Remote Sensing Symposium}, 2023, pp. 297--300.

\bibitem{10170600}
R.~Nagaraj, V.~Arulvadivelan, K.~Gouthamkumar, K.~Dharshen, and L.S. Kumar,
\newblock ``{Surface water mapping and volume estimation of Lake Victoria using Machine Learning Algorithms},''
\newblock in {\em {2023 International Conference on Signal Processing, Computation, Electronics, Power and Telecommunication (IConSCEPT)}}, 2023, pp. 1--6.

\bibitem{inproceedings}
H.~Mazumdar, M.P. Murphy, S.~Bhatkande, H.P. Emerson, D.I. Kaplan, and H.A. Gohel,
\newblock ``Optimized machine learning model for predicting groundwater contamination,''
\newblock in {\em 2022 IEEE MetroCon}. IEEE, 2022, pp. 1--3.

\bibitem{rs11060617}
S.~Hafeez, M.S. Wong, H.C. Ho, M.~Nazeer, J.~Nichol, D.~Abbas, S.and~Tang, K.H. Lee, and L.~Pun,
\newblock ``{Comparison of Machine Learning Algorithms for Retrieval of Water Quality Indicators in Case-II Waters: A Case Study of Hong Kong},''
\newblock {\em Remote Sensing}, vol. 11, no. 6, 2019.

\end{thebibliography}
\end{document}